\documentclass[10pt,twocolumn,letterpaper]{article}

\usepackage{iccv}
\usepackage{times}
\usepackage{epsfig}
\usepackage{graphicx}
\usepackage{amsmath}
\usepackage{amssymb}

\usepackage[pagebackref=true,breaklinks=true,letterpaper=true,colorlinks,bookmarks=false]{hyperref}

\iccvfinalcopy 

\def\iccvPaperID{****} 

\ificcvfinal\fi
\begin{document}

\title{Relation-Aware Pyramid Network (RapNet) for temporal action proposal   : \\ Submission to ActivityNet Challenge 2019}

\author{Jialin Gao\\
Cooperative Medianet Innovation Center\\
Shanghai Jiao Tong University\\
{\tt\small jialin\_gao@sjtu.edu.cn}
\and
Zhixiang Shi, Jiani Li, Yufeng Yuan, Jiwei Li, Xi Zhou\\
CloudWalk Technology Co., Ltd\\
{\tt\small \{shizhixiang, lijiani, yuanyufeng, lijiwei, zhouxi\}@cloudwalk.cn}
}

\maketitle

\begin{abstract}
   In this technical report, we describe our solution to \textbf{temporal action proposal (task 1)} in \textbf{ActivityNet Challenge 2019}. First, we fine-tune a ResNet-50-C3D CNN on ActivityNet v1.3 based on Kinetics pretrained model to extract snippet-level video representations and then we design a Relation-Aware Pyramid Network (RapNet) to generate temporal multiscale proposals with confidence score. After that, we employ a two-stage snippet-level boundary adjustment scheme to re-rank the order of generated proposals. Ensemble methods are also been used to improve the performance of our solution, which helps us achieve 2nd place.
\end{abstract}

\section{Video Representation}
For temporal action proposal task in ActivityNet Challenge 2019, we employ the two-stage strategy, which requires encoding visual content of video and generating proposal sequentially. In this section, we describe how to obtain video representions for proposal generation. 

First, we extract RGB frames from the original video and then adopt an Residual C3D network \cite{he2016deep, tran2015learning} to encode the visual context. In order to obtain compact features, we compose snippets sequence $\Omega = \{\omega\}_{i =1}^{T'} $ of a given video, where $T'$ is the rescaled temporal length of snippets and each snippet $\omega_i$ with $L$ frames. Our Residual C3D network takes these snippets as input to generate encoded representations of a given video.

For details, we show the backbone C3D network in Tab.\ref{tb.1} and we use ResNet-50 and ResNet-101 for this task. The net is first trained on Kinetics \cite{carreira2017quo} with ImageNet \cite{deng2009imagenet} pretrain weights provided in Pytorch Model Zoo, which got 71.48\% top-1 accuracy and then fine-tuned on ActivityNet-v1.3 dataset. Finally, we add another fully connected layer to generate 256-dimension vector for each snippet so that a video can be represented as a $T'\times 256$ feature map. Due to the different temporal length between videos, we resize them to a fixed temporal length so that every video representation is $128\times 256$ for proposasl generation.

\begin{table}[t]
	\centering
	\begin{tabular}{c|c|c}
		\hline
		Layer name & Net Architecture & Output size \\
		\hline
		$conv_{1}$ & $1\times7\times7$, 64, s(1, 2, 2) & $L\times112\times112$ \\
		\hline
		$maxpool_{1}$ & $1\times3\times3$, s(1, 2, 2) & $L\times56\times56$ \\
		\hline
		$res_{2}$ & 
		$\begin{matrix}
		1 \times 1 \times 1 & 64 \\
		1 \times 3 \times 3 & 64 \\
		1 \times 1 \times 1 & 256
		\end{matrix} \times 3$
		& $L\times56\times56$ \\
		\hline
		$maxpool_{2}$ & $2\times1\times1$, s(2, 1, 1) & $L/2\times56\times56$ \\
		\hline
		$res_{3}$ &
		$\begin{matrix}
		1 \times 1 \times 1 & 128 \\
		1 \times 3 \times 3 & 128 \\
		1 \times 1 \times 1 & 512
		\end{matrix} \times 4$
		& $L/2\times28\times28$ \\
		\hline
		$res_{4}$ &
		$\begin{matrix}
		3 \times 1 \times 1 & 256 \\
		1 \times 3 \times 3 & 256 \\
		1 \times 1 \times 1 & 1024
		\end{matrix} \times k$
		& $L/2\times14\times14$ \\
		\hline
		$res_{5}$ &
		$\begin{matrix}
		3 \times 1 \times 1 & 512 \\
		1 \times 3 \times 3 & 512 \\
		1 \times 1 \times 1 & 2048
		\end{matrix} \times 3$
		& $L/2\times7\times7$ \\
		\hline
		$pool_{3}$ & global average pool & $1\times1\times1$ \\
		\hline
	\end{tabular}
	\caption{Our backbone ResNet-50 C3D model, when $k=6$.  The input with $L\times224\times224$ dimensions was performed downsampling in the temporal size only at layer $maxpool_2$. $T\times H\times W$ represents the dimensions on time, height, weight of filter kernel size and output feature maps.}
	\label{tb.1}
\end{table}

\begin{figure*}[ht]
	\centering
	\includegraphics[scale=0.5]{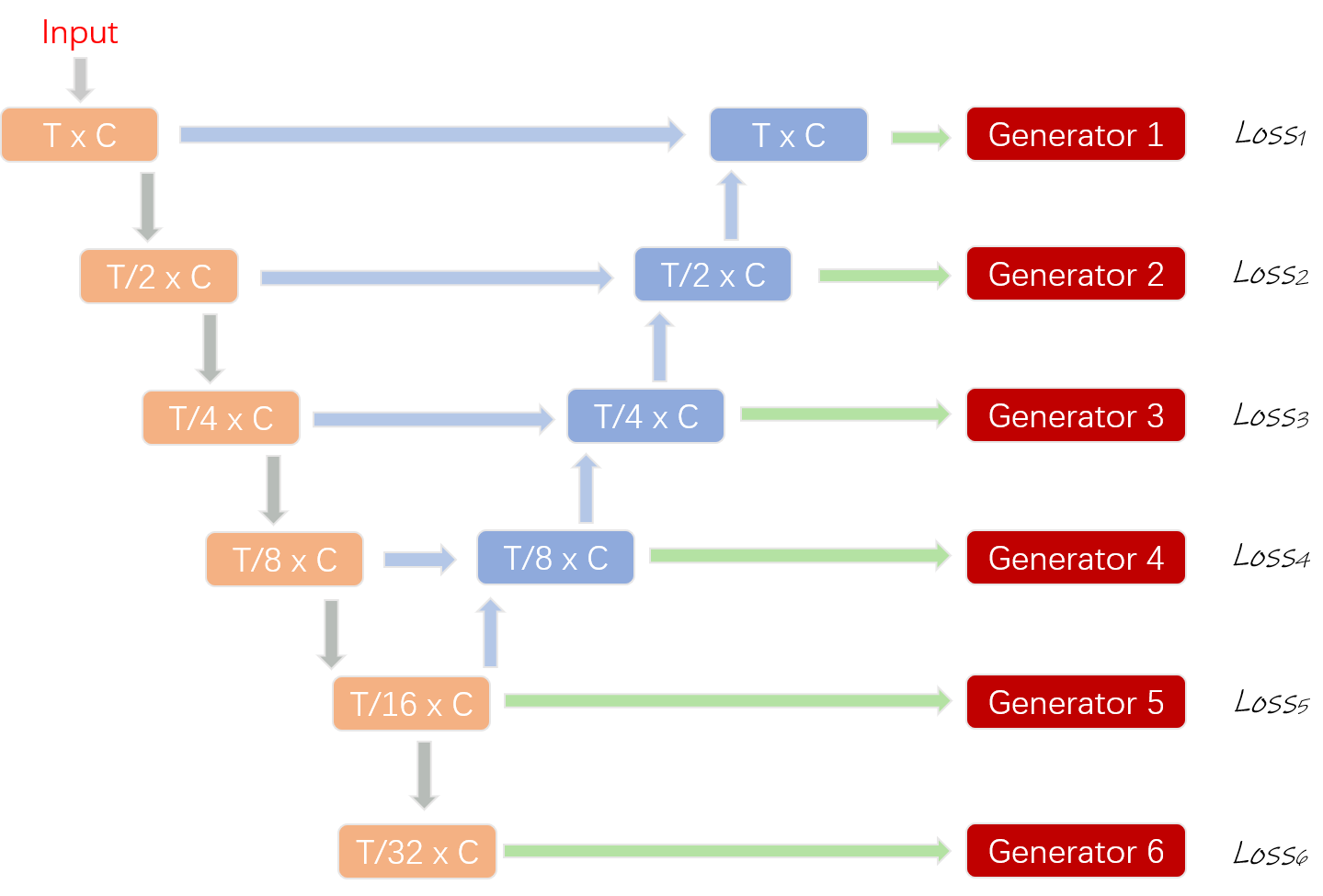}
	\caption{The architecture of our action proposal generation network. The orange blocks and blue ones are enhanced by self-attention modules and FPN respectively.}
	\label{retinanet}
\end{figure*}
\section{Action Proposal Generation}
In this section, we describe our Relation-Aware Pyramid Network (RapNet) for temporal proposal generation based on the above video representations. We employ anchor based method and design a temporal pyramid network enhanced with self-attention module \cite{vaswani2017attention} to generate multi-scale proposals. First, we use K-means to cluster a certain number of anchors, similar to YOLO \cite{redmon2018yolov3}. Then we generate multi-scale proposals based on these anchors via our RapNet, which is shown in Fig.\ref{retinanet}.

\textbf{Pre-defined anchors:} According to the annotation, we employ K-means algorithm to select anchor boxes. In experiment, we find 12 anchors achieve the best performance, which indicates every generators has 2 anchor boxes.

\textbf{Multiscale proposal generator:} In order to enlarge the receptive field of temporal convolution and capture the long-range dependency, we apply self-attention \cite{vaswani2017attention} module to take into the relationship between snippets on top-down process and use 1D FPN \cite{lin2017feature} on bottom-up to generate multi-scale proposals with 6 generators. This design can help these features in different scales capturing the long-range contextual information for better representation of visual content.

\textbf{Label assignment.} For anchor-based proposal prediction, we tag a binary label for each anchor instance, similar to YOLO \cite{redmon2018yolov3}, that a positive label is assigned to the one with highest Interaction-over-Union (IoU) with corresponding ground-truth instance, otherwise negative. In this way, a ground truth instance only can match one anchor so that the boundary regression and IoU loss only consider positive samples. Due to this label assignment, it is imbalanced that the ratio between positive and negative training samples. Thus we adopt a screening strategy to ignore some negative instances for confidence loss. A negative instance will be ignored if the highest IoU overlap between ground-truth instances with all proposal predictions is larger than a threshold $\theta_{iou}$.

\textbf{Loss function:} We use these loss functions: objectness loss, regression function, and IOU loss, which share the same definition as previous works. Hence, we optimize the following loss function $L_{prop}$:
\begin{equation}
\centering
\begin{aligned}
\lambda_{conf} [ \frac{1}{N_{pos}} \sum_{i=0}^{N-1}\sum_{j=0}^{\frac{T}{2^i}}\sum_{k=0}^{M}\Delta^{ins}_{ijk} f_{conf}(\hat{p}^{jk}_{conf},{p}^{jk}_{conf} ) \\
+ \frac{1}{N_{neg}} \sum_{i=0}^{N-1}\sum_{j=0}^{\frac{T}{2^i}}\sum_{k=0}^{M}\Lambda^{ins}_{ijk} f_{conf}(\hat{p}^{jk}_{conf},{p}^{jk}_{conf} )] \\
+ \lambda_c \frac{1}{N_{pos}} \sum_{i=0}^{N-1}\sum_{j=0}^{\frac{T}{2^i}}\sum_{k=0}^{M}\Delta^{ins}_{ijk} f_c(\hat{p}^{jk}_c, p^{jk}_c) \\
+ \lambda_w \frac{1}{N_{pos}} \sum_{i=0}^{N-1}\sum_{j=0}^{\frac{T}{2^i}}\sum_{k=0}^{M}\Delta^{ins}_{ijk} f_w(\hat{p}^{jk}_w, p^{jk}_w) \\
+ \lambda_{iou} \frac{1}{N_{pos}} \sum_{i=0}^{N-1}\sum_{j=0}^{\frac{T}{2^i}}\sum_{k=0}^{M}\Delta^{ins}_{ijk} (1 - iou_{jk})
\end{aligned}
\label{prop_loss}
\end{equation}
where $N_{pos}$ and $N_{neg}$ represent the number of positive $\Delta^{ins}_{ijk}$ and screened negative $\Lambda^{ins}_{ijk}$ training instances respectively while ${p}^{jk}_{conf}$ is the binary label in label assignment, $f_{conf}(\cdot)$ and $f_c(\cdot)$ are binary cross-entropy with logits loss functions. $f_w(\cdot)$ represents smooth-$L_1$ loss and $iou_{jk}$ is the interaction-over-union between a prediction with ground truth. We set $\lambda_{conf}=0.2$ and other weights as 1 for training RapNet.

\section{Boundary Adjustment}
Due to the start and end point of an activity is usually not very clear even if considering the global context, we degisn a two-stage strategy to adjust the boundary of the proposals generated via our RAPNet above. First we employ the refined PEM module in BSN \cite{lin2018bsn} with frames actioness to refine the boundary of proposals. And then we suppres the redundant proposals with soft-NMS \cite{bodla2017soft} to obtain reordering results. Finally, with the frames actioness, we can apply watershed algorithm in TAG \cite{xiong2017pursuit} to adjust proposals' boundary again. The improvement of these method is shown in Tab.\ref{compare}.

Our single model APG can achieve 69.61\% AUC and PEM module can improve the results by 0.74\%, TAG refinement can still increase the AUC to 70.65\%, which indicates our boundary adjustmnet is very effective for temporal action proposal.

\begin{table}
	\centering
	\small
	\begin{tabular}{c|c|c|c|c|c}
		\hline\hline
		Method & AR@1 & AR@5 & AR@10 & AR@100 & AUC(val) \\
		\hline\hline
		APG & 34.70\% & 49.65\%& 57.23\%& 78.21\%& 69.61\%\\
		\hline
		+PEM & 35.21\%& 50.29\%& 58.17\%& 78.74\%& 70.35\%\\
		\hline
		+TAG & 35.30\%& 50.44\%& 58.31\%& 79.15\%& 70.65\%\\
		\hline
	\end{tabular}
	\caption{Comparison between our RAPNet with boundary adjustment schemes on ActivityNet validation in terms of AR@AN and AUC}
	\label{compare}
\end{table}

\section{Ensemble}
In order to enhance the performance of our solution, we introduce several improvements as following:
\begin{itemize}
	\item {\it Video representations:} we use ResNet-50 C3D and ResNet-101 C3D to encode the visual content of video.
	\item  {\it Anchor boxes:} we use different anchor boxes, namely anchor 12 and 18, to generate different proposals
\end{itemize}

After ensemble, we achieves 71.51\% on the validation set and 71.38\% on the testing server.

\section{Conclusion}
In this work, we propose a novel action proposal generation network enhanced with self-attention module and FPN, called Relation-Aware Pyramid Network (RAPNet), for temporal proposal task. We also introduce a two-stage scheme to refine the boundary of proposals to improve the performance.

{\small
\bibliographystyle{ieee}
\bibliography{egbib}

\begin{thebibliography}{10}\itemsep=-1pt

\bibitem{bodla2017soft}
N.~Bodla, B.~Singh, R.~Chellappa, and L.~S. Davis.
\newblock Soft-nms--improving object detection with one line of code.
\newblock In {\em Proceedings of the IEEE International Conference on Computer
  Vision}, pages 5561--5569, 2017.

\bibitem{carreira2017quo}
J.~Carreira and A.~Zisserman.
\newblock Quo vadis, action recognition? a new model and the kinetics dataset.
\newblock In {\em proceedings of the IEEE Conference on Computer Vision and
  Pattern Recognition}, pages 6299--6308, 2017.

\bibitem{deng2009imagenet}
J.~Deng, W.~Dong, R.~Socher, L.-J. Li, K.~Li, and L.~Fei-Fei.
\newblock Imagenet: A large-scale hierarchical image database.
\newblock 2009.

\bibitem{he2016deep}
K.~He, X.~Zhang, S.~Ren, and J.~Sun.
\newblock Deep residual learning for image recognition.
\newblock In {\em Proceedings of the IEEE conference on computer vision and
  pattern recognition}, pages 770--778, 2016.

\bibitem{lin2018bsn}
T.~Lin, X.~Zhao, H.~Su, C.~Wang, and M.~Yang.
\newblock Bsn: Boundary sensitive network for temporal action proposal
  generation.
\newblock In {\em Proceedings of the European Conference on Computer Vision
  (ECCV)}, pages 3--19, 2018.

\bibitem{lin2017feature}
T.-Y. Lin, P.~Doll{\'a}r, R.~Girshick, K.~He, B.~Hariharan, and S.~Belongie.
\newblock Feature pyramid networks for object detection.
\newblock In {\em Proceedings of the IEEE Conference on Computer Vision and
  Pattern Recognition}, pages 2117--2125, 2017.

\bibitem{redmon2018yolov3}
J.~Redmon and A.~Farhadi.
\newblock Yolov3: An incremental improvement.
\newblock {\em arXiv preprint arXiv:1804.02767}, 2018.

\bibitem{tran2015learning}
D.~Tran, L.~Bourdev, R.~Fergus, L.~Torresani, and M.~Paluri.
\newblock Learning spatiotemporal features with 3d convolutional networks.
\newblock In {\em Proceedings of the IEEE international conference on computer
  vision}, pages 4489--4497, 2015.

\bibitem{vaswani2017attention}
A.~Vaswani, N.~Shazeer, N.~Parmar, J.~Uszkoreit, L.~Jones, A.~N. Gomez,
  {\L}.~Kaiser, and I.~Polosukhin.
\newblock Attention is all you need.
\newblock In {\em Advances in neural information processing systems}, pages
  5998--6008, 2017.

\bibitem{xiong2017pursuit}
Y.~Xiong, Y.~Zhao, L.~Wang, D.~Lin, and X.~Tang.
\newblock A pursuit of temporal accuracy in general activity detection.
\newblock {\em arXiv preprint arXiv:1703.02716}, 2017.

\end{thebibliography}
}

\end{document}